\documentclass[journal]{IEEEtran}
\usepackage{url,hyperref,lineno,microtype,subcaption}
\usepackage{cite}
\usepackage{amsmath,amssymb,amsfonts}
\usepackage[utf8]{inputenc}
\usepackage{mathtools}
\usepackage{algorithmic}
\usepackage[ruled,vlined]{algorithm2e}
\usepackage{graphicx}
\usepackage{textcomp}
\usepackage[inline]{enumitem}
\usepackage{xcolor}
\usepackage{multirow}
\usepackage{makecell}

\ifCLASSINFOpdf
\else
\fi
\begin{document}
%
\title{Federated Learning with Spiking Neural Networks}
%
%
%

\author{Yeshwanth~Venkatesha, 
        Youngeun~Kim, 
        Leandros~Tassiulas,
        and~Priyadarshini~Panda
        \\ Department of Electrical Engineering, Yale University, USA 
\thanks{Yeshwanth~Venkatesha, Youngeun Kim, Leandros~Tassiulas, and Priyadarshini~Panda are with the Department of Electrical Engineering, Yale University, New Haven, CT, USA 
(Corresponding author: Yeshwanth~Venkatesha, e-mail: yeshwanth.venkatesha@yale.edu).}
} 

\maketitle

\begin{abstract}
As neural networks get widespread adoption in resource-constrained embedded devices, there is a growing need for low-power neural systems. Spiking Neural Networks (SNNs) are emerging to be an energy-efficient alternative to the traditional Artificial Neural Networks (ANNs) which are known to be computationally intensive. 
From an application perspective, as federated learning involves multiple energy-constrained devices, there is a huge scope to leverage energy efficiency provided by SNNs.
Despite its importance, 
there has been little attention on training SNNs on a large-scale distributed system like federated learning.
In this paper, we bring SNNs to a more realistic federated learning scenario.
Specifically, we propose a federated learning framework for decentralized and privacy preserving training of SNNs.
To validate the proposed federated learning framework, we experimentally evaluate the advantages of SNNs on various aspects of federated learning with CIFAR10 and CIFAR100 benchmarks.
We observe that SNNs outperform ANNs in terms of overall accuracy by over $15\%$ when the data is distributed across a large number of clients in the federation while providing up to $5.3\times$ energy efficiency.
In addition to efficiency, we also analyze the sensitivity of the proposed federated SNN framework to data distribution among the clients, stragglers, and gradient noise and perform a comprehensive comparison with ANNs. 
The source code is available at \href{https://github.com/Intelligent-Computing-Lab-Yale/FedSNN}{https://github.com/Intelligent-Computing-Lab-Yale/FedSNN}.

\end{abstract}

\begin{IEEEkeywords}
Neuromorphic Computing, Federated Learning,  Spiking Neural Network, Energy-efficient Deep Learning, Event-based Processing
\end{IEEEkeywords}

%
\IEEEpeerreviewmaketitle

\section{Introduction}
\IEEEPARstart{N}{euromorphic}  learning is being explored as a resilient low-power alternative to conventional Deep Learning in the perspective of both algorithm and hardware \cite{roy2019towards, neuromorphic_review_signal_processing}. With the release of neuromorphic chips like IBM TrueNorth \cite{IBM_TrueNorth} and Intel Loihi \cite{intel_loihi}, we are not very far from neuromorphic processors becoming a part of everyday life. A class of Neuromorphic Learning algorithms called Spiking Neural Networks (SNNs) have gained widespread attention for their ability to achieve comparable performance to that of deep neural networks at considerably less computation cost and extreme energy efficiency \cite{snn_imagenet_first, deep_snn}. SNNs, analogous to the electrical activity in the human brain, utilize the discrete spike mechanism with Integrate-and-Fire (IF) or Leaky-Integrate-and-Fire (LIF) neuron units to transmit information.

Considering their energy efficiency, SNNs have a huge potential to be deployed in applications on embedded devices. However, these models need to be continuously trained and updated according to the new data collected at these devices. Conventional cloud-based machine learning collects data from the devices, transfers it to a central location, and trains the model offline. However, owing to increasing privacy concerns, it is unwise to transfer data out of the device. Modern collaborative on-device learning methods such as Federated Learning offers a solution for this.
Federated Learning \cite{fl_google, konevcny2016federated, gboard, fed_application, application_wireless, shlezinger2020uveqfed} is used for private and data-secure distributed learning among mobile devices (or clients) and a cloud server. This enables model training without the transfer of data from the client \cite{fl_google} to the cloud. This is particularly helpful when the data is sensitive and privacy is of paramount importance. Federated learning has been deployed in day-to-day applications such as Google Gboard \cite{gboard} and is likely to be deployed in multiple practical applications in healthcare, retail, finance, wireless communication, etc \cite{fed_application, application_wireless}. It provides the advantage of continuously learning and updating the model as and when the new data is available which is more practical compared to offline training. It can also enable user personalization as we train models locally on user data \cite{fallah2020personalized}. 

\begin{figure*}[h]
  \begin{center}
    \includegraphics[width=0.85\textwidth]{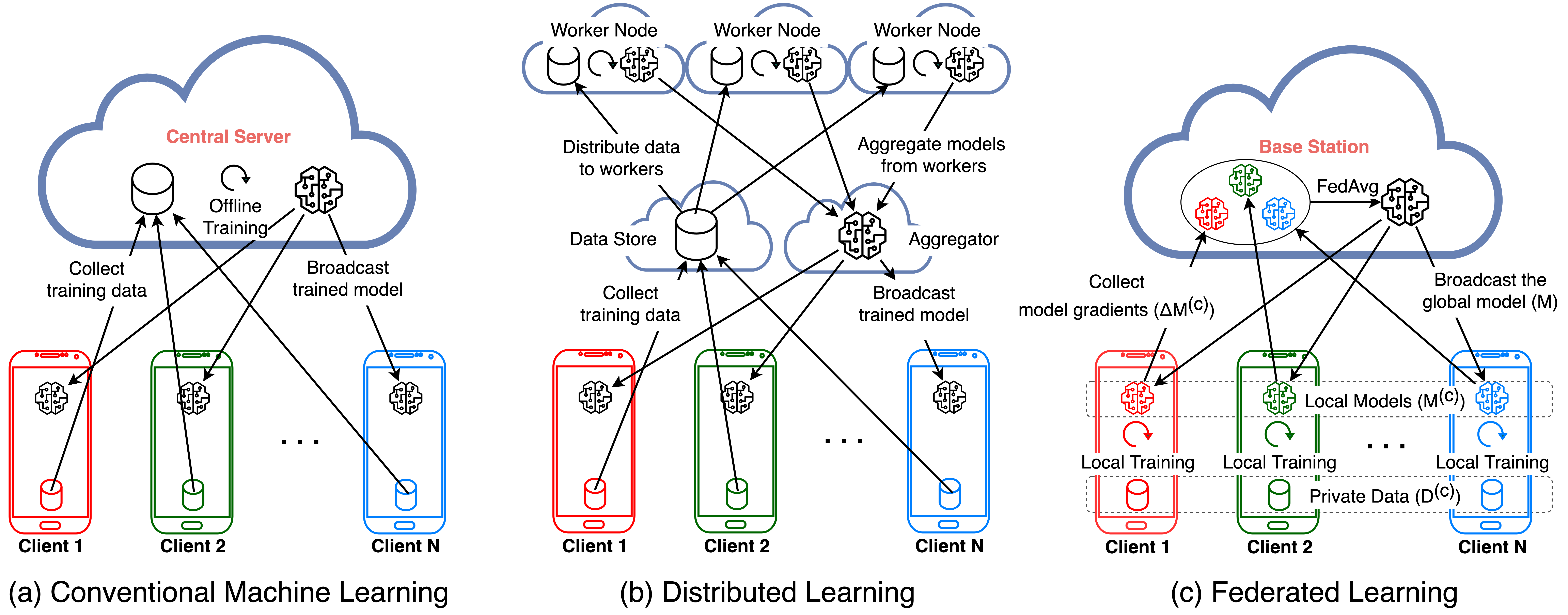}
  \end{center}
  \caption{Schematic diagram to illustrate distributed learning and federated learning. (a) Conventional Learning --- the data is collected from the clients and the model is trained in an offline manner. (b) Distributed Learning --- the dataset is collected at a single data store and then it is distributed across multiple worker nodes for training. (c) Federated Learning --- In this case, the training is performed at the client, and model gradients are communicated to the base station which aggregates the gradients and updates the global model, and broadcasts it back to all the clients. Each of these processes is iterative and models at the clients are periodically updated. }
   \label{fig:fedlearning_concept}
\end{figure*}

A major distinguishing factor of federated learning over traditional distributed training lies in the properties of the data distribution in the clients. Fig. \ref{fig:fedlearning_concept} summarizes the difference between conventional machine learning, distributed learning and federated learning. In conventional machine learning, the data is collected into a single location and a model is trained offline on a single machine. However, as modern systems deal with enormous amount of data it is not always feasible to train a model on a single machine. Hence, distributed learning aims to accelerate the training process by parallelizing the training across multiple workers by distributing the training dataset. Here, since all the data is available at one location (i.e. cloud server), it can be Independent and Identically Distributed (IID) among the workers. This results in each worker having nearly the same amount of data and equal distribution of all the classes resulting in a stable learning method and faster convergence. On the other hand, in federated learning, the clients are not allowed to share their local data to the cloud server. The clients can only share their local model updates with the server. Furthermore, owing to the heterogeneity in the client devices, the IID assumption is not valid in the case of federated learning. Handling the non-IID data among the clients is one of the key aspects of federated learning that is widely studied \cite{zhao2018federated, sattler2019robust, wang2020optimizing}. This problem is further amplified as the number of clients increases. 

Another important factor to consider while designing a federated learning system is the impact of stragglers \cite{li2020federated, smith2017federated}. As the devices are not controlled by the central server and typically the updates happen over an unreliable network, it is not guaranteed that all the clients will send their updates to the server within an acceptable time limit. Hence, the server might end up with fewer updates than expected. Further, recent studies have shown that some of the data can be recovered from the model updates \cite{wang2019beyond, zhu2020deep}. To counter this data leakage, there is a body of work involving differential privacy \cite{dwork2008differential} which obfuscates the gradients of the model with noise \cite{wei2020federated}. Hence, the training algorithm for federated learning must be resilient to noise in the model updates communicated to the server. 

A majority of existing SNN literature examines SNNs only through the lens of energy efficiency and is limited to standard machine learning settings where the entire dataset is available in one place. There has been little attention on utilizing SNNs in more practical systems beyond classic image recognition task. A potential reason for this limitation is the lack of robust training algorithms for SNNs.
Traditionally, the non-linear thresholding functionality in LIF or IF spiking neurons is known to prohibit the conventional backpropagation-based training methods for SNNs. Hence, the training of SNNs was limited to unsupervised learning methods like Spike Timing Dependent Plasticity \cite{stdp} and conversion of fully trained ANN to SNN \cite{diehl2015fast, diehl2016conversion}. However, recent findings illustrate methods to reliably train SNNs using an approximate gradient backpropagation method. This method unrolls the network and performs a Backpropagation Through Time (BPTT) similar to that of recurrent neural networks \cite{neftci2019surrogate, snn_imagenet_first}. In addition to this, a recent work uses batch normalization in the time dimension---Batch Normalization Through Time (BNTT) in order to achieve high performing SNNs by training from scratch \cite{bntt}. Such gradient-based training methods enable SNNs to be used in distributed training methods such as federated Learning. Further, recent studies suggest that SNNs are robust to adversarial attacks \cite{adversarial_snn,kim2021visual}. As security is a major concern in federated learning applications, we believe SNNs can be a viable alternative to ANNs \cite{mothukuri2021survey}. 

In contrast to previous works which are limited to extracting computation and energy efficiency with the spiking model, we look at SNNs from the perspective of robust distributed training. 
We propose a method to train SNNs in a federated learning paradigm and conduct a series of structured experiments to study the robustness of SNNs trained in a federated manner with respect to the different aspects of federated learning.
Previously, the authors of \cite{skatchkovsky2020federated} propose an on-device training method for SNNs in a federated setting using a probabilistic SNN model. In contrast to their work, we use a mainstream gradient-based approach instead of a probabilistic model to make the training faster and more scalable. Moreover, since the reported experiments in \cite{skatchkovsky2020federated} were performed on the MNIST-DVS dataset, it may not be straightforward to extend it to mainstream computer vision problems. To the best of our knowledge, our work is one of the first to train SNNs with federated learning on complex image datasets and perform a comprehensive analysis of robustness, performance, and energy efficiency.

\textbf{Contributions:} 
In this paper, we design a federated learning framework to train low-power SNNs in a distributed and privacy preserving manner.
We experimentally evaluate the feasibility of federated learning with SNNs for standard image classification tasks with CIFAR10 and CIFAR100 datasets. We highlight that, as the number of devices in the federation increases, SNNs outperform ANNs by $>15\%$ on CIFAR10 and $>25\%$ on CIFAR100 with VGG9 model. We study the effect of skewness in the class distribution among the clients on the performance of the system. We evaluate and compare the performance of SNNs to ANN counterparts with respect to robustness to stragglers and robustness to gradient noise and finally, provide estimated energy efficiency of SNNs compared to ANNs.

\section{Background}
In this section, we provide a brief background on federated learning, spiking neural networks and the  batch normalization technique for SNNs, called Batch Normalization Through Time (BNTT).

\subsection{Federated Learning}

A typical federated learning system consists of a set of $N$ clients and a base station. The base station broadcasts the initial model $M_0$ to all the clients to start the training process. The initial model can be a model trained offline on a public dataset. Each of the clients $c = 1, 2, ..., N$, maintain their own private dataset $D^{(c)}$ and locally train a model by iterating over the local dataset $D^{(c)}$ to obtain a locally trained model at client $c$ (say $M^{(c)}$). The size of the data retained in the client depends on the storage and computation capacity of the device. The model update from each client is the accumulated gradients over the course of local training. These updates are periodically communicated to the base station for aggregation. Each iteration of communication of the model updates from clients to the server is defined as a federated learning \textit{round}. At \textit{round} $r$, the gradients $\Delta M^{(c)}_{r}$  from a subset of participating clients $P_{r} (\ll N)$ is sent to the base station. The participating clients can be selected at random or based on a defined scheduling algorithm according to the needs of the specific application. The base station performs a weighted average of the gradients to obtain the updated global model as shown in Eqn. \ref{fl_equation}.
\begin{equation}\label{fl_equation}
M_{r + 1} = M_{r} + \frac{1}{\sum_{c \epsilon P_{r}} |D^{(c)}_{r}|}\sum_{c \epsilon P_{r}}|D^{(c)}_{r}|\Delta M^{(c)}_{r},
\end{equation}
where $|D^{(c)}_{r}|$ denotes the number of data samples used for local training in client $c$ at round $r$. This aggregation of the model updates from the clients is described as $FedAvg$ algorithm in \cite{fl_google}. There have been multiple enhancements to the $FedAvg$ algorithm such as $FedProx$, that uses a regularization term to stabilize federated training \cite{FedProx} and $FedMA$, which performs a layer-wise matching and averaging of the models to handle heterogeneous data \cite{fedma}. For simplicity, in this article, we shall be using only $FedAvg$. The process of federated learning is illustrated in Fig. \ref{fig:fedlearning_concept}.

\subsection{Spiking Neural Networks}

\begin{figure}[t]
  \begin{center}
    \includegraphics[width=\columnwidth]{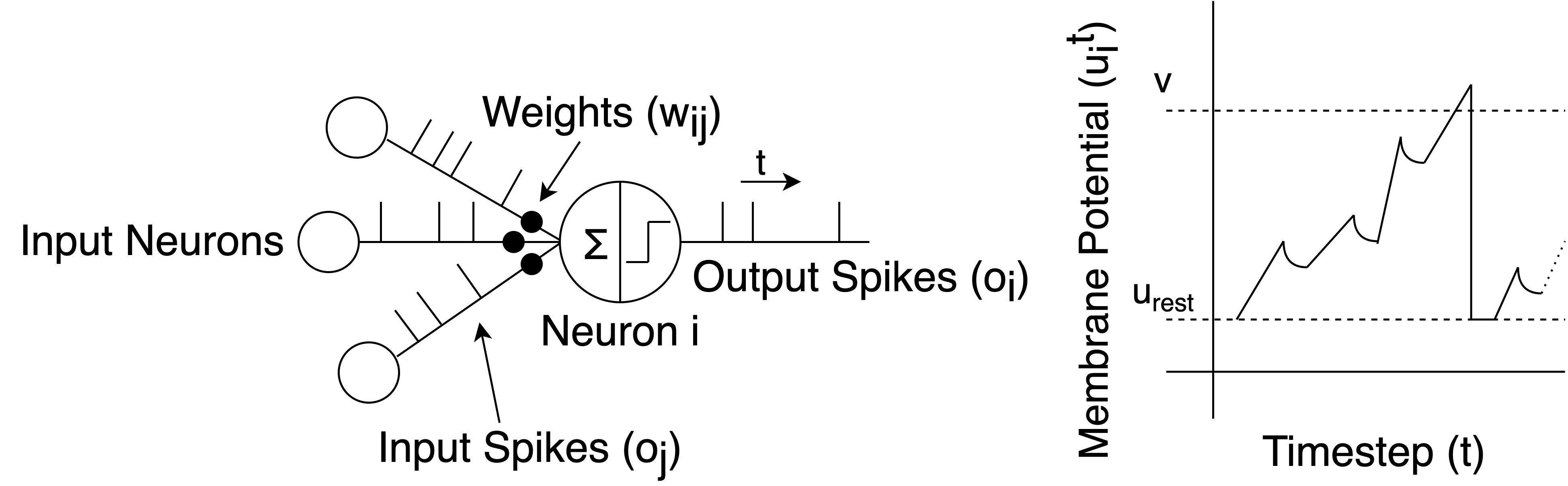}
  \end{center}
  \caption{
  The illustration of leaky-integrate-and-fire behavior of a spiking neuron.
  }
   \label{fig:snn_concept}
\end{figure}

A spiking neuron is modeled by an Integrate-and-Fire (IF) mechanism --- each neuron keeps track of its membrane potential by accumulating the incoming spikes scaled by their corresponding synaptic weights and generates a spike when its membrane potential reaches a certain threshold. In the Leaky-Integrate-and-Fire (LIF) variant, the membrane potential also leaks at a constant rate. As illustrated in Fig. \ref{fig:snn_concept}, for a given neuron $i$ with a set of input neurons $N$, the incoming spikes from the input neurons are weighted by the parameters $w_{ij}$ for all $j \in N$ and accumulated as the membrane potential of the neuron. Once the membrane potential reaches a certain threshold $v$, it generates an output spike. 
Following a spike, the membrane potential is reset to the resting potential $u_{rest}$ or in case of a soft reset, it is reduced by the threshold value. This process is repeated for $T$ timesteps.
In discrete form, the LIF mechanism is modeled as follows \cite{discrete_lif,rathi2020enabling}: 
\begin{equation}\label{discrete_lif_eqn}
\begin{split}
    u^{t}_{i} = \lambda u^{t - 1}_{i} + \sum_{j \in N} w_{ij} o^{t-1}_{j}, \quad \quad \quad \\
    \text{where} \quad \lambda < 1 \quad \text{and} \quad
    o^{t - 1}_{i} = \begin{cases} 1 &\mbox{if } u^{t - 1}_{i} > v \\
0 & \mbox{otherwise } \end{cases}.
\end{split}
\end{equation}
Here, $o^{t}_{i}$ is the binary output that takes the value $1$ if the neuron $i$ produces a spike at timestep $t$ and $0$ otherwise, $u^{t}_{i}$ represents the membrane potential of neuron $i$ at timestep $t$, $\lambda$ is a constant factor by which membrane potential reduces every timestep, $v$ is the threshold, and $N$ specifies the set of input neurons connected to neuron $i$ with $w$ specifying their corresponding weights. 

The operation of a spiking neuron can be unrolled along the time dimension similar to a recurrent neural network (RNN) \cite{neftci2019surrogate}. Hence the gradients are accumulated over $T$ timesteps and can be calculated as follows:
\begin{equation}\label{snn_backprop}
    \Delta w_{ij} = \sum_{t = 1}^{T} \frac{\partial \mathcal{L}}{\partial w_{ij}^{t}} = \sum_{t = 1}^{T} \frac{\partial \mathcal{L}}{\partial o_{i}^{t}} \frac{\partial o_{i}^{t}}{\partial u_{i}^{t}} \frac{\partial u_{i}^{t}}{\partial w_{ij}^{t}},
\end{equation}
where, $\mathcal{L}$ is the loss function being optimized. In case of image classification, categorical cross entropy loss is widely used.
However, as $o_i^t$ is a thresholding function, its' derivative is a Dirac-Delta function that is not defined at the time of spike and zero everywhere else. Thus, computing $\Delta w_{ij}$ is intractable. To overcome this, the derivative of threshold function is approximated with several surrogate functions such as piece-wise linear function and the exponential function \cite{neftci2019surrogate, snn_bp}. The surrogate gradient descent method using a piece-wise linear approximation is defined as:
\begin{equation}\label{surrogate_gd}
    \frac{\partial o_i^t}{\partial u_i^t} = \xi \max \{0, 1-  \ | \frac{u_i^t - v}{v} \ | \},
\end{equation}
where, $\xi$ is a decay factor for back-propagated gradients and $v$ is the threshold value.
The hyperparameter $\xi$ should be set based on the total number of timesteps $T$. As gradients are accumulated at every time step, it is recommended to use a smaller $\xi$ for large $T$ to avoid exploding gradient problem.

\subsection{Training Spiking Neural Networks with BNTT}

The authors of \cite{bntt} propose a method to leverage batch normalization in temporal dimension to improve the training performance of SNNs. With this method, it is possible to train low latency and high-performing SNNs from scratch without the need for a fully trained ANN model. This is achieved by associating a local learning parameter to each time-step and thereby expanding the batch norm layer through time.
During the forward propagation, the BNTT layer is applied after each convolutional/linear layer as:
\begin{equation}
    \begin{split}
    u_i^t = & \lambda u_i^{t-1} + \textup{BNTT}_{\gamma^t}(\sum_{j \epsilon N} w_{ij}o^t_j ) \\ 
    = & \lambda u_i^{t-1} + \gamma_i^t (\frac{\sum_{j \epsilon N} w_{ij}o^t_j  - \mu^t_i}{\sqrt{(\sigma_i^t)^2 + \epsilon}}),
    \end{split}
    \label{eq:LIFwithBN}
\end{equation}
where $\mu_{i}^t$ and $\sigma_{i}^t$ are the mean and variance from the samples in a mini-batch $\mathcal{B}$ at time step $t$.
A global mean and variance is calculated by computing an exponential moving average over training iterations which are used to normalize the validation data at inference. The parameter $\gamma$ is  learnt using backpropagation and different $\gamma^t$ values are used for each time-step $t$ for efficient inference.


\begin{figure}[t]
  \begin{center}
    \includegraphics[width=\columnwidth]{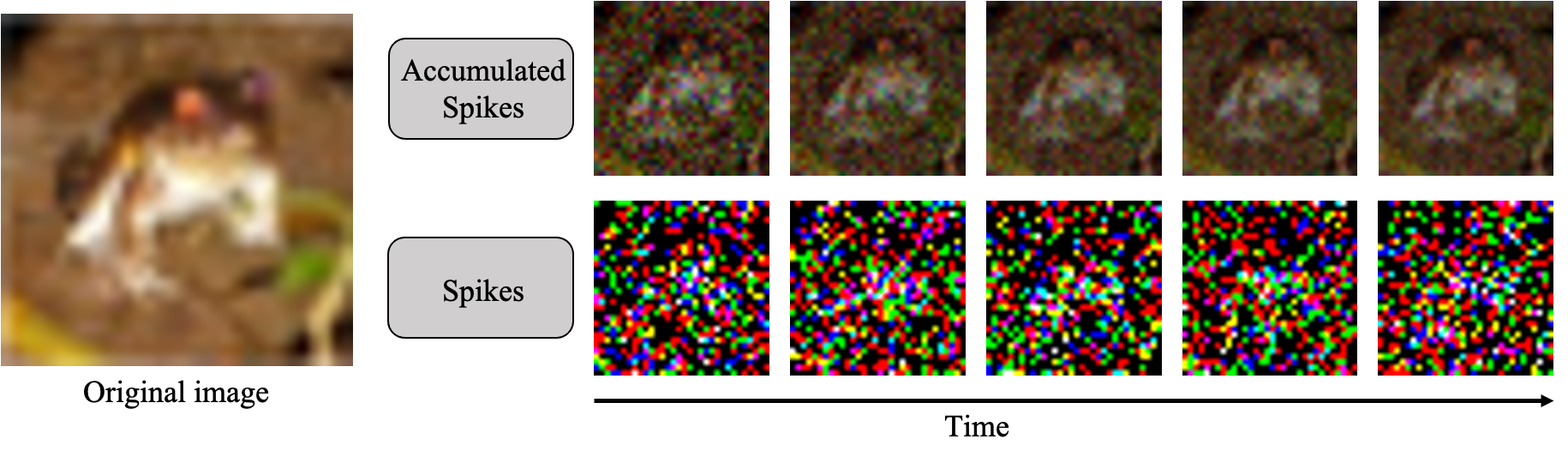}
  \end{center}
  \caption{ Example of rate coding. As time goes on, the accumulated spikes represent similar image to original image. We use an image from the CIFAR10 dataset.
  }
  \label{fig:possion_coding}
\end{figure}

\section{Federated SNN Training}

Given a base station and a set of $N$ clients with each client $c$ having its own private local dataset $D^{(c)}$, we train a single SNN model without transferring the dataset out of the client. The base station initially broadcasts a randomly initialized SNN model to each of the $N$ clients. As described in Algorithm \ref{algo:fed_snn}, each client trains a local copy of the model $M^{(c)}$ in parallel using the BNTT training technique. The BNTT training process (described in procedure $BNTT{\text -}Training$ in Algorithm \ref{algo:fed_snn}) starts by encoding the pixel values into spike trains of length $T$ using Poisson rate coding. In Poisson coding, the number of spikes generated is matched with the corresponding pixel intensity. At each timestep, a random number is sampled between the minimum and maximum possible pixel intensity ($I_{min}$, $I_{max}$), and a spike is generated if this random number is less than the real pixel intensity. Hence, the spike generated at each timestep corresponding to a pixel is stochastic while the total number of spikes generated over the timesteps is proportional to the pixel intensity. In Fig. \ref{fig:possion_coding}, we illustrate the process of spike encoding using poisson generator. Note that, if we accumulate spikes over time, we get an image similar to the original.

For a given timestep $t=1,2,...,T$, for each layer $l=1,2,...,L-1$ with weight tensor $W_l$, BNTT parameter $\gamma_l$ and threshold $v_l$, we perform forward propagation to update the set of membrane potentials $U^t_l$ and when the membrane potentials exceed the layer threshold $v_l$, output spikes $O^t_l$ are generated:
\begin{equation}\label{snn_fwd}
\begin{split}
    U_l^{t} = \lambda U_l^{t-1} + \textup{BNTT}_{\gamma^t_l}(W_{l}, O^{t-1}_{l-1}), \quad \quad \quad \\
    \text{where} \quad \lambda < 1 \quad \text{and} \quad
    O_l^{t} = \begin{cases} 1 &\mbox{if } U^{t - 1}_{l} > v_l, \\
0 & \mbox{otherwise } \end{cases} 
\end{split}
\end{equation}
Note that we use upper case notation to denote vectors and lower case to denote scalar values. Here, $U_l^t$ and $O_l^t$ is the set of membrane potentials and output spikes respectively of all the neurons of layer $l$ at timestep $t$. $W_l$ is the set of weights $w_{ij}$ connecting neurons of layer $l - 1$ to neurons of layer $l$. $\lambda$ is the constant leak factor of membrane potential.
The membrane potential at the last layer $L$ is accumulated without the leak factor (i.e. $\lambda = 1$ in Eqn. \ref{snn_fwd}) to make the output values continuous which are passed through a softmax layer to get the model predictions. 

\begin{algorithm}[t]
  \SetAlgoLined\DontPrintSemicolon
  \SetKwInOut{Input}{Input}
  \SetKwInOut{Output}{Output}
  \SetKwInOut{Hyperparameters}{Hyperparams}
  \SetKwFunction{algo}{Fed-SNN}\SetKwFunction{proc}{BNTT-Training}
  \SetKwProg{myalg}{Algorithm}{}{}
  \Input{Set of $N$ clients with Local Datasets $D^{(c)} \forall c = 1, 2, ..., N$}
  \Output{Trained Model $M$ }
  \Hyperparameters{Number of rounds (R), Number of local epochs (E), Number of timesteps (T)}
  \myalg{\algo{}}{
   Initialize Model $M$ with random weights\;
  \For{round $r\gets0$ \KwTo $R$}{
   Broadcast the current model $M$ to all clients\;
    \For{Client $c\gets0$ \KwTo $N$ (in parallel)}{
      \For{$local\_epoch\gets0$ \KwTo $K$}{
        perform \proc{} on $M_r^{(c)}$. \;
        
        }
    }
    Randomly select $P$ participating devices\;
    Aggregate the gradients (Eqn. \ref{fl_equation}).\;
  }
   }
   \KwRet\;{}
  \setcounter{AlgoLine}{0}
  \SetKwProg{myproc}{Procedure}{}{}
  \myproc{\proc{}}{
    \ForEach{mini-batch B}{
         \For{time step $t\gets0$ \KwTo $T$}{
             $O \leftarrow PoissonGenerator(B)$\;
             \For{Layer $l\gets1$ \KwTo $L-1$}{
                Forward propagate (Eqn. \ref{snn_fwd}).\;
             }
             Accumulate membrane potential at the last layer.\;
            }
            Compute Loss (Eqn. \ref{eq:loss_fn}) and Backpropagate the model gradient ${\delta}M^{(c)}$. \;
        }
   }
   \KwRet\;
  \caption{Federated SNN Training Method}\label{algo:fed_snn}
\end{algorithm}

\begin{figure*}[t]
  \begin{center}
    \includegraphics[width=0.8\textwidth]{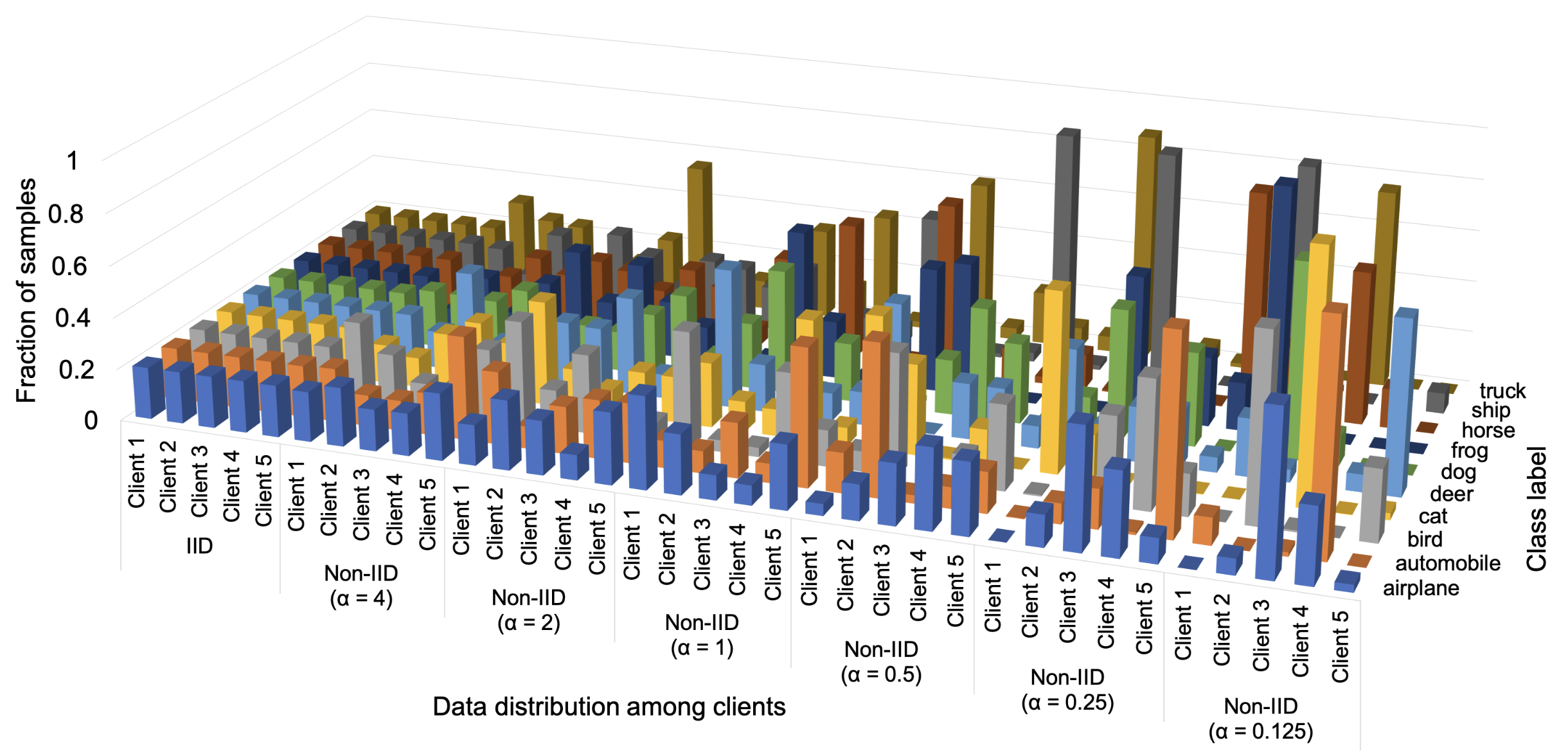}
  \end{center}
  \caption{Distribution of CIFAR10 dataset among 5 clients for different degrees of non-IID. In the case of IID, the 10 classes are equally distributed among all the clients with each client getting 1/5 of the samples of each class. As we increase the parameter alpha to the Dirichlet Distribution, the class composition at each client becomes more varied. The extreme case with $\alpha = 0.125$ results in no examples of certain classes in some clients. }
   \label{fig:non_iid_example}
\end{figure*}

The model gradient $\delta M^{(c)}$ at client $c$ is calculated backpropagating the loss $\mathcal{L}$ through the unrolled network by iterating over the local dataset $D^{(c)}$ according to Eqn. \ref{snn_backprop} and \ref{surrogate_gd}. For the case of image classification, the loss $\mathcal{L}$ is defined by categorical cross-entropy function  between the accumulated membrane potentials of the last layer $U^T_L$ and the ground truth labels $Y$ as follows:
\begin{equation}\label{eq:loss_fn}
    \mathcal{L}(U_L^T, Y) = -log (\frac{exp({U^T_L[Y]})}{\sum_{j}exp(U^T_L[j])}),
\end{equation}
The BNTT training process updates each of the local client model $M^{(c)}$ at epoch $e$ as:
\begin{equation}
    M^{(c)}_{e} = M^{(c)}_{e - 1} - {\eta}{\delta}M^{(c)}_e,
\end{equation}
where $\eta$ is the local learning rate and ${\delta}M^{(c)}_e$ is the model gradients of client $c$ at epoch $e$. 
A full pass through the local dataset constitutes a \textit{local epoch} and the gradients accumulated over $K$ local epochs forms the aggregated model update $\Delta{M^{(c)}} = \sum_{e=1}^{K}\eta\delta M^{(c)}$. The number of \textit{local epochs} $K$ is set depending on the volume of data and the available compute resources on each client. 
After every $K$ \textit{local epochs}, a randomly chosen subset of $P$ clients will send their aggregated model updates $\Delta{M^{(i)}}$ to the base station which in turn updates the global model according to the FedAvg Algorithm (Eqn. \ref{fl_equation}). This cycle of broadcast, local model update, gradient communication, and global model update constitutes one \textit{round} of federated learning. This process is repeated for $R$ \textit{rounds} to train the model. The process is detailed in Algorithm \ref{algo:fed_snn}.
In production systems, this is a lifelong process to keep the model up to date with the latest data distribution. This is extremely helpful as many modern applications rely on quickly adapting to the latest trends. Note that instead of sending the model updates $\Delta{M^{(c)}}$ to the base station, the clients can also send the full model ${M^{(c)}}$. The base station can, then, perform an equivalent computation and get the same updated model. However, transferring gradients can result in efficient communication if a relevant gradient compression technique is used \cite{optimal_grad_compression}.


\section{Experiments}

\begin{figure*}[t]
  \begin{center}
    \includegraphics[width=0.65\textwidth]{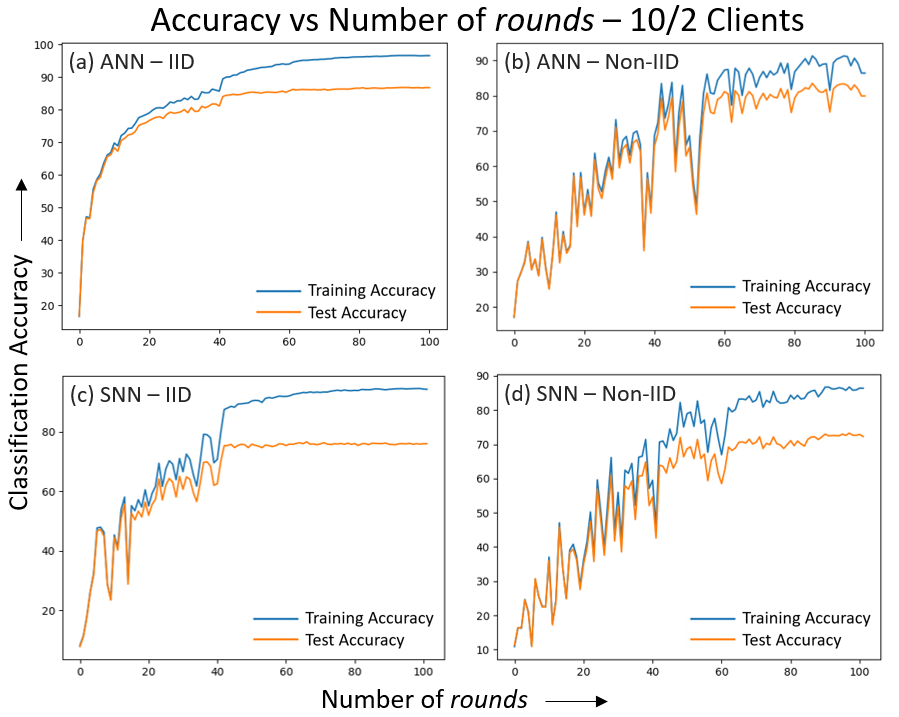}
  \end{center}
  \caption{Training progress of the VGG9 model trained on CIFAR10 dataset for 10 clients with 2 clients participating in each round. 
  We show the contrast between IID and non-IID and between SNN and ANN. While the training curve is smoother for ANN on IID data, SNN has a relatively smoother curve in the case of non-IID data distribution. Particularly, in the course of ANN training with non-IID data, we occasionally observe a sharp decline in the performance. While the overall training is not fully smooth, the training of SNNs is comparatively robust to sharp changes. } 
   \label{fig:fedlearn_acc}
\end{figure*}

To evaluate our method, we perform structured experiments with VGG9 network \cite{vgg} on CIFAR10 and CIFAR100 benchmarks \cite{CIFAR}. We use the standard training and validation datasets provided in the original dataset with 50000 32x32 RGB images of training data and 10000 images of validation data. We use two strategies to distribute the training dataset among $N$ clients:
\begin{enumerate*}[label=(\roman*)]
  \item IID $-$ each client has an equal number of samples and nearly the same proportion of all classes.
  \item non-IID $-$ sample the proportion of each class using a Dirichlet distribution with concentration parameter $\alpha = 0.5$ to obtain non-identical datasets similar to previous federated learning works \cite{bayesianfl, fedma}.
\end{enumerate*}

In Fig. \ref{fig:non_iid_example}, we illustrate the difference between IID and non-IID and describe how we control the $ \alpha$ parameter in Dirichlet distribution to generate different extent of non-IID distribution.
Although non-IID is more realistic, we use the IID strategy to establish a baseline to compare against. 
For simplicity, we assume the local datasets do not change between rounds, i.e. $D^{(i)}_r = D^{(i)} $ $\forall r$.
The validation set is held out and is used to evaluate all the models. We use the validation accuracy on this held-out dataset as the performance metric throughout our experiments. All the reported validation accuracy values are calculated by averaging across three repetitions of the experiment.
We use a batch size of 32 across all the experiments. We train locally on each client for 5 epochs which is termed as \textit{local epochs} and train the global model for 100 \textit{rounds}. We consider different number of total and participating clients to study how the algorithm scales with the number of devices. We use the convention of $N/P$ to specify the client split where $N$ denotes the total number of clients and $P$ denotes the number of participating devices in each round.

We use VGG9 as the base model for our experiments. VGG9 consists of 7 convolutional layers and two fully connected layers with pooling layers to reduce the dimension. Traditionally max pooling is used in ANN literature and average pooling is more common in the SNN literature. To make the comparison simple, we use average pooling for both ANN and SNN.  We use SGD optimizer with an initial learning rate of 0.001 and weight decay of 5e-4 for ANN while we use SGD with an initial learning rate of 0.1 and momentum of 0.95 for SNN. The learning rate is reduced by a factor of 5 after 40, 60, and 80 epochs. Since the training mechanisms are different, naturally, one set of hyperparameters will not be optimal for both ANN and SNN. 

In Fig. \ref{fig:fedlearn_acc}, we plot the training curve for one instance of federated learning with 10 total clients and 2 clients participating in each round to investigate the training progress. We observe a smoother curve with IID data distribution compared to non-IID in both ANN and SNN. This is expected since the non-IID case has a high variance in the data among the devices. In the case of ANN with non-IID data, we observe that the accuracy drops sharply at some of the rounds (round 38 and 52 in Fig. \ref{fig:fedlearn_acc}b) implying that ANNs are sensitive to every model update whereas in SNN training there is no such sharp performance drops. This implies that federated learning with SNNs is more robust to abrupt deviations in the model updates.

In subsequent sections, we study in detail, the performance of our method on various facets of federated learning. In Section \ref{section:scalability}, we evaluate the performance of the model as the number of clients increases. In Section \ref{section:participating_clients}, we study the effect of the number of participating clients in each round.
In Section \ref{section:niid_study}, we expand on the details of non-IID and study the impact of skewness in the data distribution on the performance of the model. In Section \ref{section:stragglers} and Section \ref{section:gradient_noise} we investigate if there is a significant drop in accuracy with respect to having stragglers and noise in gradients respectively. Finally, in Section \ref{section:energy_estimation}, we estimate the energy consumption of SNNs and compare it to that of ANNs.

\subsection{Scalability}\label{section:scalability}

\begin{table*}[t]
\caption{Final validation accuracy reached by the models after 100 rounds for different numbers of total and participating clients on CIFAR10 and CIFAR100. The key observation to note is that the performance deterioration is steeper in the case of ANN compared to that of SNN as N/P increases.}\label{tab:result_summary}
\begin{center}
\begin{tabular}{|c|c|c|c|c|c|c|c|c|}
\hline
 & \multicolumn{4}{|c|}{\textbf{CIFAR10}} & \multicolumn{4}{|c|}{\textbf{CIFAR100}} \\
 \cline{2-9} 
{\textbf{Clients}} & \multicolumn{2}{|c|}{\textbf{ANN}} & \multicolumn{2}{|c|}{\textbf{SNN}} & \multicolumn{2}{|c|}{\textbf{ANN}} & \multicolumn{2}{|c|}{\textbf{SNN}} \\
\cline{2-9} 
                    (N/P)          & \textbf{\textit{IID}}       & \textbf{\textit{non-IID}}     & \textbf{\textit{IID}}       & \textbf{\textit{non-IID}} & \textbf{\textit{IID}}       & \textbf{\textit{non-IID}}     & \textbf{\textit{IID}}       & \textbf{\textit{non-IID}}    \\
\hline
1/1                   & 87.27         & 87.27            & 77.54      & 77.54   & 59.98 & 59.98 & 46.84 & 46.84   \\
5/5                 & 85.40         & 83.15            & 76.91     & 74.54  & 58.39 & 53.55 & 48.54 & 43.49 \\
10/2                & 82.81         & 79.68          & 76.44     & 73.94   & 55.56 & 53.55 & 47.25 & 41.00    \\
20/5              & 78.25        & 73.02           & 75.01     & 68.80     & 47.47 & 44.80 & 49.95 & 46.64  \\
100/10            & 50.84         & 44.33                 & 67.54     & 58.71  & 12.29 & 13.12 & 42.79 & 40.91 \\ 

150/15            & 42.82         & 36.86                 & 63.85     & 59.32  & 8.25 & 8.39 & 36.61 & 37.35 \\ 

200/20            & 36.37         & 33.39                 & 58.76     & 55.31  & 4.61 & 5.08 & 32.52 & 32.13 \\ 
\hline
\end{tabular}
\label{table_results}
\end{center}
\end{table*}

\begin{figure*}[t]
  \begin{center}
    \includegraphics[width=0.8\textwidth]{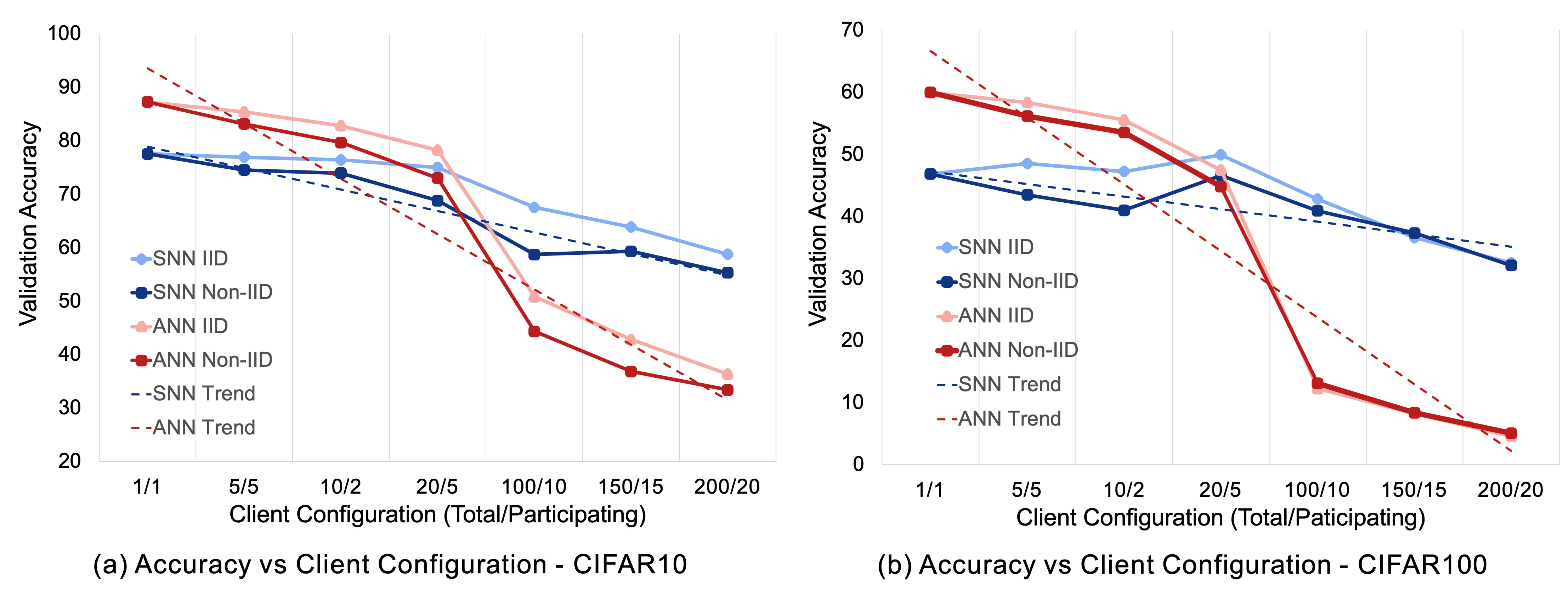}
  \end{center}
  \caption{Performance trend of ANN vs SNN for VGG9 model trained on CIFAR10 and CIFAR100 as we increase the number of clients. We observe a similar trend on both the datasets where ANNs perform better than SNNs when the number of clients is low but as the number of clients is scaled up, the performance of ANNs plummet. }
   \label{fig:acc_trend}
\end{figure*}

Since real-world federated systems involve a huge number of devices, a federated learning model must be scalable with the number of devices. To evaluate the scalability of our method, we perform a series of experiments steadily increasing the number of total and participating clients in the federated learning process (Fig. \ref{fig:acc_trend} and Table \ref{tab:result_summary}). We report the performance on several combinations of clients for both CIFAR10 and CIFAR100 datasets in Table \ref{tab:result_summary} and provide plots in Fig. \ref{fig:acc_trend} to visualize the trend. 
While the ANN achieves superior accuracy in the baseline case of 1/1 (N/P) configuration, the accuracy drops sharply as the number of clients increases. On the other hand, the performance degradation is not as steep in the case of SNN. SNNs surpass the performance of ANNs when the total number of clients is increased to 100. This phenomenon is more pronounced in the case of CIFAR100 (shown in Fig. \ref{fig:acc_trend}b) where the ANN model accuracy goes below 20\% for 100/10 clients which is practically unusable. Therefore, as the number of clients increase, we observe that SNNs outperform the corresponding ANNs.

\subsection{Impact of Number of Participating Clients}\label{section:participating_clients}

\begin{figure*}[t]
  \begin{center}
    \includegraphics[width=0.8\textwidth]{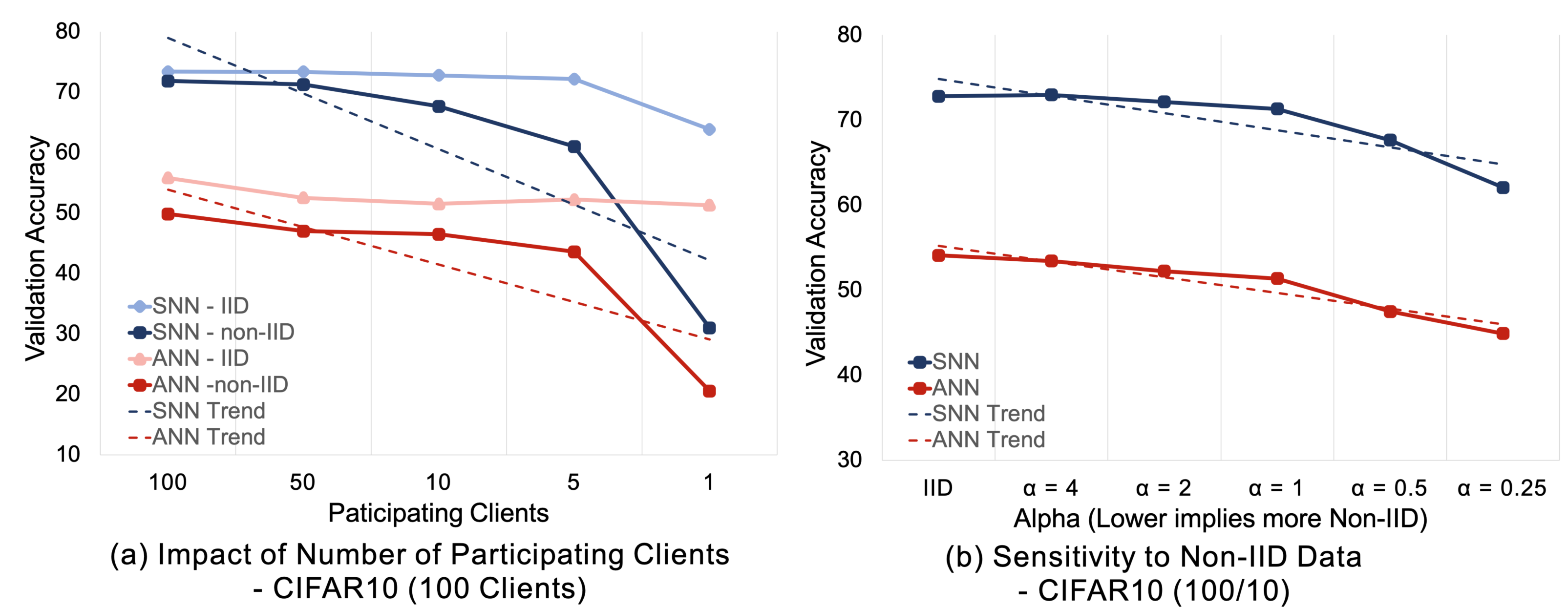}
  \end{center}
  \caption{ (a) Impact of number of clients participating in each round. For the case of CIFAR10 with 100 clients, we decrease the total participating devices from 100 to 1 and study the performance. We observe a similar trend with both ANNs and SNNs. (b) Effect of having more skewed class distribution among the clients. We increase the non-IID nature of the data distribution and observe the validation accuracy of the final model. We use the case with CIFAR10 divided among 100 clients and 10 clients participating in each round. We observe a similar trend of decreasing performance as the distribution becomes more non-IID in case of both ANNs and SNNs.}
   \label{fig:niid_analysis}
\end{figure*}

While total number of clients is one aspect of scalability, the fraction of clients participating in each round also plays an important role in the performance of a federated learning system. Especially in the case of non-IID, it is desirable to have a model that can generalize well with less number of clients participating in each round. To study the impact of number of participating clients, we use the case with CIFAR10 trained with 100 clients and progressively decrease the number of participating clients from 100 to 1. We observe, in Fig. \ref{fig:niid_analysis}a, that there is a gradual performance drop as the number of participating clients decrease in case of non-IID in both ANNs and SNNs. In case of IID, since all the clients contain similar data distribution, both ANNs and SNNs are preserving the performance as the number participating clients reduce. This implies that SNNs are robust in generalizing the model using updates from only a fraction of the total clients. 

\subsection{Sensitivity to Data Distribution}\label{section:niid_study}
In addition to scalability with the number of devices, the nature of data distribution across the devices is another factor that impacts the performance of a federated learning system. In this section, we study how SNNs perform as the distribution of classes among the clients is varied. As the non-IID data distribution is synthetically generated by sampling from Dirichlet distribution, we can control the skewness by varying the parameter $\alpha$ of the Dirichlet distribution \cite{bayesianfl, fedma}. Fig. \ref{fig:non_iid_example} illustrates how the fraction of samples of different classes are distributed among different clients. As the value of $\alpha$ decreases, the class composition gets more skewed. This skewness will be further amplified as the number of clients increase. In Fig. \ref{fig:niid_analysis}b, we consider the models trained on CIFAR10 with 100/10 clients at different extent of non-IID. We start with $\alpha = 4$ and progressively reduce it by the factor of $2$ as long as the model does not diverge and observe the performance of the system. We note that there is a steady decline in the performance of both ANNs and SNNs as the data becomes more non-IID. This implies that SNNs are equally robust in handling non-IID data as compared to ANNs. 


\subsection{Sensitivity to Stragglers}\label{section:stragglers}

\begin{figure*}[t]
  \begin{center}
    \includegraphics[width=0.8\textwidth]{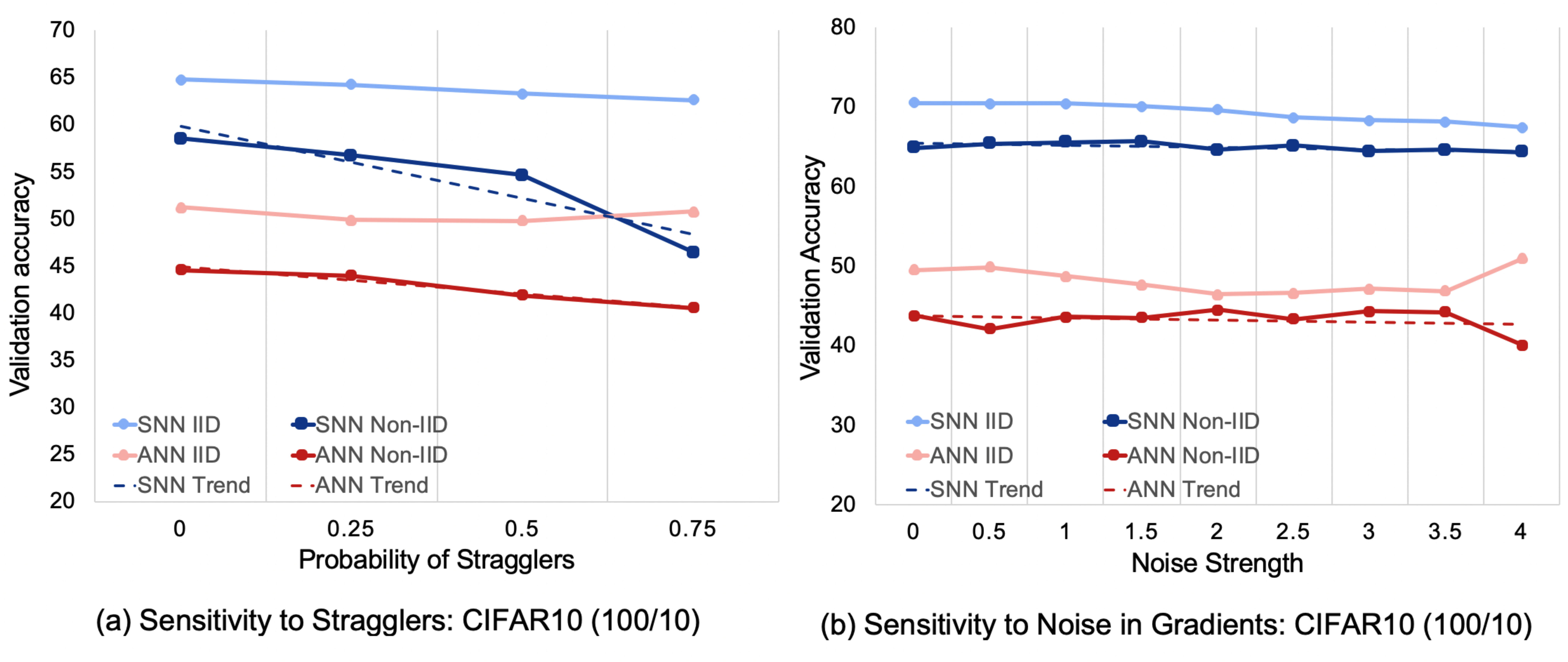}
  \end{center}
  \caption{(a) Impact of stragglers on the performance. We observe training SNNs with our method is robust to stragglers. There is a noticeable drop in performance in extreme cases. However, 75\% straggler probability is extreme. (b) Impact of noise in the gradients. Both SNNs and ANNs show equal robustness with added noise to the gradients.}
   \label{fig:straggler_and_noise_analysis}
\end{figure*}

Since real-world federated learning applications involve training with millions of devices, it is impractical to assume the communication of gradients will be successful from all the selected devices. Hence, the model needs to be robust to handle devices failing to communicate the gradients. These devices are referred to as stragglers \cite{li2020federated, smith2017federated}. In this section, we analyze the impact of stragglers on the performance of the final SNN model. We use the case with 100 total clients and 10 participating clients for a VGG9 model trained on CIFAR10 data. We evaluate the model performance by varying the probability of a device failing to communicate the model updates to the server. To consider a \textit{round} to be successful, at least one out of $P$ participating devices has to communicate its update to the base station. Hence, one out of $P$ participating clients is guaranteed to send the update, while the rest of the clients drop out with a probability value. We consider different levels of probabilities and plot the trend in Fig. \ref{fig:straggler_and_noise_analysis}a. Since the updates are similar across the clients in the case of IID, the impact of the missing updates is minimal resulting in a nearly flat curve. However, the impact of stragglers is prominent in the case of non-IID data for both ANN and SNN. We observe a similar trend between ANN and SNN up to a probability value of $0.5$. The steep drop in performance of SNNs in the extreme case of stragglers (for probabilities $>0.5$) suggests that SNNs are more sensitive to stragglers when the data distribution is non-IID. However, it is worth noting that SNN is still performing better than the corresponding ANN in presence of stragglers. Hence, we conclude SNNs are robust in handling stragglers to a considerable extent on par with the corresponding ANNs on CIFAR10 dataset.


\subsection{Sensitivity to Noise in Gradients}\label{section:gradient_noise}


\begin{figure*}[t]
  \begin{center}
    \includegraphics[width=0.6\textwidth]{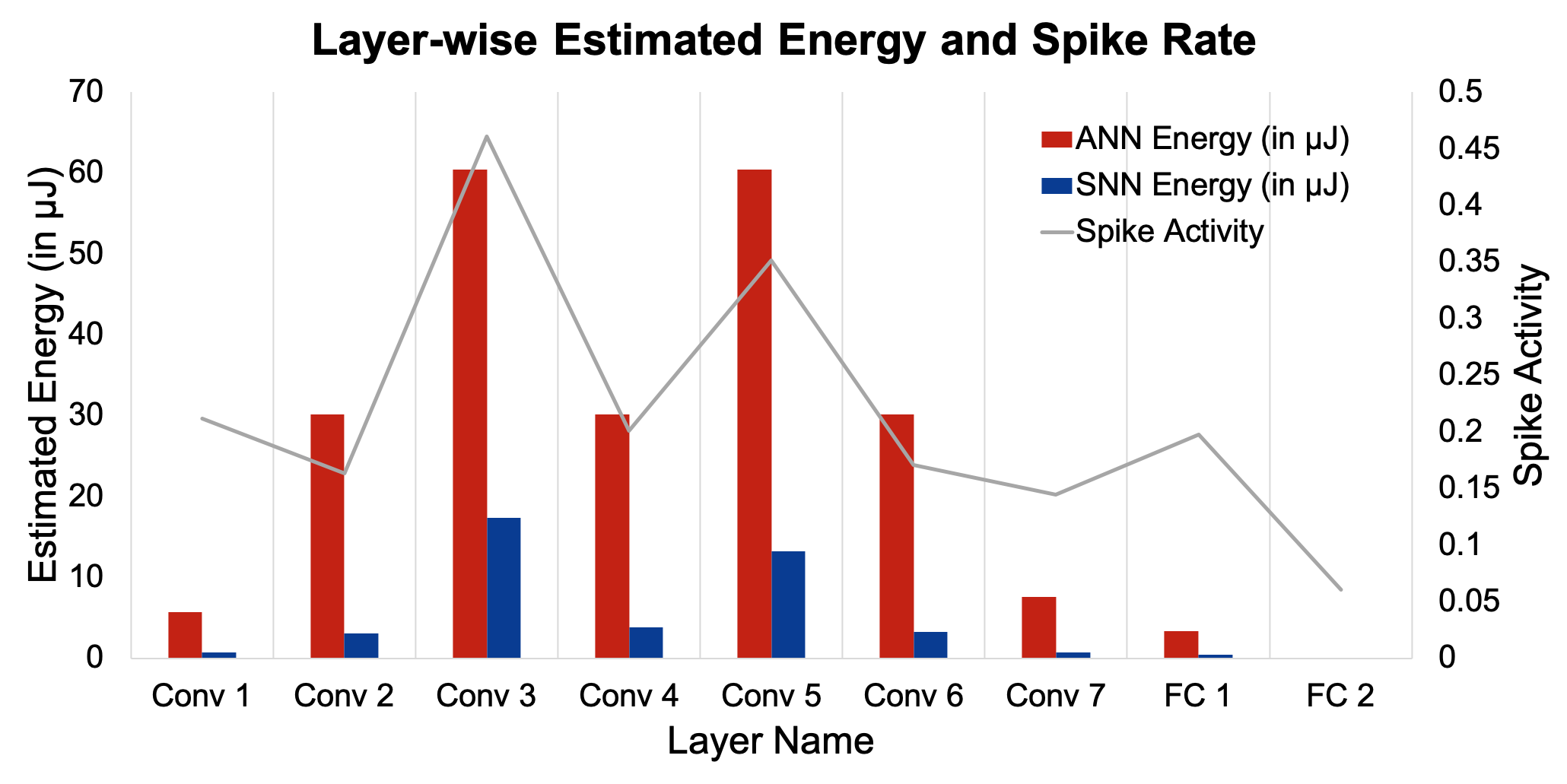}
  \end{center}
  \caption{Estimated inference energy across layers for SNN and ANN VGG9 model trained on CIFAR10 dataset. We see that SNNs are considerably more energy efficient compared to ANNs.}
   \label{fig:energy_plot}
\end{figure*}

Straightforward federated learning is not enough to preserve data privacy as there are methods by which some of the data and its attributes can be recovered from the gradients communicated between the clients and the server. Hence, the gradients are often obfuscated with added noise before sending to the base station. In this section, we evaluate the performance of the federated model with respect to the added noise in the gradients. As in the previous section, we use the case with 100 total clients and 10 participating clients for a model trained on CIFAR10 data. We add gaussian noise $N(0,1)$ multiplied by noise strength. The noise strength is increased up to 4 implying an added noise with a mean of $0$ and standard deviation of $4$. The observation is captured in Fig. \ref{fig:straggler_and_noise_analysis}b. For the IID case, in both SNN and ANN, we observe a slight impact of noise as the performance is gradually decreasing. On the contrary, there is no visible impact on the performance in the case of non-IID resulting in a nearly flat line. This can be explained by the fact that gradient updates from non-IID is inherently diverse. Hence, adding noise to the gradients has minimal effect on the performance implying that differential privacy techniques \cite{wei2020federated} can be seamlessly applied to SNNs to make the system more secure.

\subsection{Energy Estimation}\label{section:energy_estimation}

\begin{table}[t]
    \begin{center}
    \caption{Energy estimation for multiply and accumulate operations.}
    \begin{tabular}{|c|c|}
    \hline
    \textbf{Operation} & \textbf{Estimated Energy (pJ)} \\
    \hline
    32-bit Multiply ($E_{Mult}$) & 3.1\\
    32-bit Add ($E_{Add}$) & 0.1 \\
    \makecell{32-bit Multiply and Accumulate \\ ($E_{MAC} = E_{Mult} + E_{Add}$)} & 3.2 \\
    32-bit Accumulate ($E_{AC}$) & 0.1\\
    \hline
    \end{tabular}
    \label{tab:energy_calc}
    \end{center}
\end{table}

We estimate the energy consumption for 32-bit integer arithmetic operations on traditional hardware based on a 45nm CMOS process using energy calculations described in \cite{horowitz20141}. Note that, this is rather a rough estimate as we are considering only Multiply and Accumulate (MAC) operations and neglecting memory and any peripheral circuit energy. The energy for multiply and accumulate operations are summarized in Table \ref{tab:energy_calc}.
For a $k \times k$ convolution layer with $I$ input channels, $O$ output channels operating on an input feature map size $N \times N$ and resulting in output feature map of size $M \times M$ number of operations ($OPS$) is given by: $OPS = M^2 \times I \times k^2 \times O$. For a fully connected layer with $I$ inputs giving $O$ outputs, the number of operations ($OPS$) is given by $OPS = I \times O$.
Since SNNs operate with binary spikes the MAC operation reduces to an accumulate (AC) operation which leads to significant energy efficiency. The energy consumption of ANN is straight-forward $E_{ANN} = OPS \times E_{MAC}$. We calculate the energy consumption of SNN by multiplying $OPS$ with spiking rate (activity) $R$ across total timesteps, $T$ \textit{i.e.}, $E_{SNN} = OPS \times R \times T \times E_{AC}$. 
In our experiments, we computed the average spiking rate by performing inference on the entire validation set. Fig. \ref{fig:energy_plot} shows the estimated energy for each layer of ANN and SNN with VGG9 model trained on the CIFAR10 dataset for 10/2 clients with non-IID data distribution. We also observe the spike rate of the corresponding layers for the final model after 100 rounds. The total estimated energy by ANN is $\mathbf{227.99 \mu J}$ while that of SNN is $\mathbf{42.59 \mu J}$ which is $\mathbf{5.3\times}$ more efficient. The energy for ANN is constant for all the instances while that of SNN varies for each instance depending upon the spike activity.

\section{Conclusion and Discussion}
We perform a comprehensive study on the feasibility of training spiking neural networks in a federated learning paradigm. We show how to train SNNs from scratch in a federated setting using the BNTT method. We experimentally evaluate the performance and estimated energy of the VGG9 SNN trained on CIFAR10 and CIFAR100 datasets and compare them with corresponding ANNs. We study different facets of federated learning such as data distribution across the clients, stragglers, and gradient noise to find potential strengths and weaknesses of SNNs. We find that SNNs outperform ANNs when the data is distributed across a large number of clients and are equally robust to ANNs in handling non-IID data, stragglers, and noise in the gradients. We conclude that SNNs are a viable alternative to ANNs in large-scale federated learning with embedded devices, owing to their energy efficiency and superior performance when the number of clients is scaled up. 

From the series of experiments we identify that while SNNs are equally robust to ANNs in all the facets of federated learning, the key advantage of SNNs is the ability to preserve the accuracy at scale. In our experiments, as the size of the dataset is fixed, dividing it among a large number of clients results in each client having fewer samples. Naturally, the updates from these clients will overfit to their local dataset. Hence, it is difficult to aggregate these local models into a robust global model. 
Therefore, we can infer that federated training of SNNs with our method is more sample efficient i.e. the local SNN model in each client can generalize well with fewer data samples as compared to ANNs. Consequently, the aggregated global model is preserving the performance.
This phenomenon can be attributed to the fact that since SNNs contain more gradient updates due to the unrolling of the network in the time dimension, the gradients updates are more likely to become smoother providing a form of regularization to the local model. This regularization occurs intrinsically due to temporal processing in SNNs is contributing to their robustness in handling variability in the gradients from a large number of clients. 

In addition to having promising results with SNNs, this study opens up many questions to explore in future work. For example, developing a theoretical basis to explain the observation of a relatively lower decline in performance in SNNs compared to ANNs as the number of clients is increased. This would help in manifesting SNNs as a more scalable alternative to ANNs in federated learning.
Federated learning involves many more engineering intricacies such as handling delays/failures in the client updates, the communication cost of models and gradients to and from the clients, and the policy of selecting the clients in each round. SNNs can provide unique advantages in each of these directions. Exploring temporal datasets beyond static vision for federated learning with SNNs is also a potential direction to explore.




\bibliographystyle{IEEEtran}
\bibliography{main}

\begin{thebibliography}{10}
\providecommand{\url}[1]{#1}
\csname url@samestyle\endcsname
\providecommand{\newblock}{\relax}
\providecommand{\bibinfo}[2]{#2}
\providecommand{\BIBentrySTDinterwordspacing}{\spaceskip=0pt\relax}
\providecommand{\BIBentryALTinterwordstretchfactor}{4}
\providecommand{\BIBentryALTinterwordspacing}{\spaceskip=\fontdimen2\font plus
\BIBentryALTinterwordstretchfactor\fontdimen3\font minus
  \fontdimen4\font\relax}
\providecommand{\BIBforeignlanguage}[2]{{%
\expandafter\ifx\csname l@#1\endcsname\relax
\typeout{** WARNING: IEEEtran.bst: No hyphenation pattern has been}%
\typeout{** loaded for the language `#1'. Using the pattern for}%
\typeout{** the default language instead.}%
\else
\language=\csname l@#1\endcsname
\fi
#2}}
\providecommand{\BIBdecl}{\relax}
\BIBdecl

\bibitem{roy2019towards}
K.~Roy, A.~Jaiswal, and P.~Panda, ``Towards spike-based machine intelligence
  with neuromorphic computing,'' \emph{Nature}, vol. 575, no. 7784, pp.
  607--617, 2019.

\bibitem{neuromorphic_review_signal_processing}
B.~{Rajendran}, A.~{Sebastian}, M.~{Schmuker}, N.~{Srinivasa}, and
  E.~{Eleftheriou}, ``Low-power neuromorphic hardware for signal processing
  applications: A review of architectural and system-level design approaches,''
  \emph{IEEE Signal Processing Magazine}, vol.~36, no.~6, pp. 97--110, 2019.

\bibitem{IBM_TrueNorth}
\BIBentryALTinterwordspacing
P.~A. Merolla, J.~V. Arthur, R.~Alvarez-Icaza, A.~S. Cassidy, J.~Sawada,
  F.~Akopyan, B.~L. Jackson, N.~Imam, C.~Guo, Y.~Nakamura, B.~Brezzo, I.~Vo,
  S.~K. Esser, R.~Appuswamy, B.~Taba, A.~Amir, M.~D. Flickner, W.~P. Risk,
  R.~Manohar, and D.~S. Modha, ``A million spiking-neuron integrated circuit
  with a scalable communication network and interface,'' \emph{Science}, vol.
  345, no. 6197, pp. 668--673, 2014. [Online]. Available:
  \url{https://science.sciencemag.org/content/345/6197/668}
\BIBentrySTDinterwordspacing

\bibitem{intel_loihi}
M.~{Davies}, N.~{Srinivasa}, T.~{Lin}, G.~{Chinya}, Y.~{Cao}, S.~H. {Choday},
  G.~{Dimou}, P.~{Joshi}, N.~{Imam}, S.~{Jain}, Y.~{Liao}, C.~{Lin},
  A.~{Lines}, R.~{Liu}, D.~{Mathaikutty}, S.~{McCoy}, A.~{Paul}, J.~{Tse},
  G.~{Venkataramanan}, Y.~{Weng}, A.~{Wild}, Y.~{Yang}, and H.~{Wang}, ``Loihi:
  A neuromorphic manycore processor with on-chip learning,'' \emph{IEEE Micro},
  vol.~38, no.~1, pp. 82--99, 2018.

\bibitem{snn_imagenet_first}
\BIBentryALTinterwordspacing
E.~Hunsberger and C.~Eliasmith, ``Training spiking deep networks for
  neuromorphic hardware,'' \emph{CoRR}, vol. abs/1611.05141, 2016. [Online].
  Available: \url{http://arxiv.org/abs/1611.05141}
\BIBentrySTDinterwordspacing

\bibitem{deep_snn}
\BIBentryALTinterwordspacing
A.~Sengupta, Y.~Ye, R.~Wang, C.~Liu, and K.~Roy, ``Going deeper in spiking
  neural networks: Vgg and residual architectures,'' \emph{Frontiers in
  Neuroscience}, vol.~13, p.~95, 2019. [Online]. Available:
  \url{https://www.frontiersin.org/article/10.3389/fnins.2019.00095}
\BIBentrySTDinterwordspacing

\bibitem{fl_google}
\BIBentryALTinterwordspacing
H.~B. McMahan, E.~Moore, D.~Ramage, S.~Hampson, and B.~A. y~Arcas,
  ``Communication-efficient learning of deep networks from decentralized
  data,'' in \emph{Proceedings of the 20th International Conference on
  Artificial Intelligence and Statistics (AISTATS)}, 2017. [Online]. Available:
  \url{http://arxiv.org/abs/1602.05629}
\BIBentrySTDinterwordspacing

\bibitem{konevcny2016federated}
J.~Kone{\v{c}}n{\`y}, H.~B. McMahan, F.~X. Yu, P.~Richt{\'a}rik, A.~T. Suresh,
  and D.~Bacon, ``Federated learning: Strategies for improving communication
  efficiency,'' \emph{arXiv preprint arXiv:1610.05492}, 2016.

\bibitem{gboard}
\BIBentryALTinterwordspacing
A.~Hard, K.~Rao, R.~Mathews, F.~Beaufays, S.~Augenstein, H.~Eichner, C.~Kiddon,
  and D.~Ramage, ``Federated learning for mobile keyboard prediction,''
  \emph{CoRR}, vol. abs/1811.03604, 2018. [Online]. Available:
  \url{http://arxiv.org/abs/1811.03604}
\BIBentrySTDinterwordspacing

\bibitem{fed_application}
\BIBentryALTinterwordspacing
Q.~Yang, Y.~Liu, T.~Chen, and Y.~Tong, ``Federated machine learning: Concept
  and applications,'' \emph{ACM Trans. Intell. Syst. Technol.}, vol.~10, no.~2,
  Jan. 2019. [Online]. Available: \url{https://doi.org/10.1145/3298981}
\BIBentrySTDinterwordspacing

\bibitem{application_wireless}
S.~{Niknam}, H.~S. {Dhillon}, and J.~H. {Reed}, ``Federated learning for
  wireless communications: Motivation, opportunities, and challenges,''
  \emph{IEEE Communications Magazine}, vol.~58, no.~6, pp. 46--51, 2020.

\bibitem{shlezinger2020uveqfed}
N.~Shlezinger, M.~Chen, Y.~C. Eldar, H.~V. Poor, and S.~Cui, ``Uveqfed:
  Universal vector quantization for federated learning,'' \emph{IEEE
  Transactions on Signal Processing}, 2020.

\bibitem{fallah2020personalized}
A.~Fallah, A.~Mokhtari, and A.~Ozdaglar, ``Personalized federated learning: A
  meta-learning approach,'' \emph{arXiv e-prints}, pp. arXiv--2002, 2020.

\bibitem{zhao2018federated}
Y.~Zhao, M.~Li, L.~Lai, N.~Suda, D.~Civin, and V.~Chandra, ``Federated learning
  with non-iid data,'' \emph{arXiv preprint arXiv:1806.00582}, 2018.

\bibitem{sattler2019robust}
F.~Sattler, S.~Wiedemann, K.-R. M{\"u}ller, and W.~Samek, ``Robust and
  communication-efficient federated learning from non-iid data,'' \emph{IEEE
  transactions on neural networks and learning systems}, vol.~31, no.~9, pp.
  3400--3413, 2019.

\bibitem{wang2020optimizing}
H.~Wang, Z.~Kaplan, D.~Niu, and B.~Li, ``Optimizing federated learning on
  non-iid data with reinforcement learning,'' in \emph{IEEE INFOCOM 2020-IEEE
  Conference on Computer Communications}.\hskip 1em plus 0.5em minus
  0.4em\relax IEEE, 2020, pp. 1698--1707.

\bibitem{li2020federated}
T.~Li, A.~K. Sahu, A.~Talwalkar, and V.~Smith, ``Federated learning:
  Challenges, methods, and future directions,'' \emph{IEEE Signal Processing
  Magazine}, vol.~37, no.~3, pp. 50--60, 2020.

\bibitem{smith2017federated}
V.~Smith, C.-K. Chiang, M.~Sanjabi, and A.~Talwalkar, ``Federated multi-task
  learning,'' \emph{arXiv preprint arXiv:1705.10467}, 2017.

\bibitem{wang2019beyond}
Z.~Wang, M.~Song, Z.~Zhang, Y.~Song, Q.~Wang, and H.~Qi, ``Beyond inferring
  class representatives: User-level privacy leakage from federated learning,''
  in \emph{IEEE INFOCOM 2019-IEEE Conference on Computer Communications}.\hskip
  1em plus 0.5em minus 0.4em\relax IEEE, 2019, pp. 2512--2520.

\bibitem{zhu2020deep}
L.~Zhu and S.~Han, ``Deep leakage from gradients,'' in \emph{Federated
  Learning}.\hskip 1em plus 0.5em minus 0.4em\relax Springer, 2020, pp. 17--31.

\bibitem{dwork2008differential}
C.~Dwork, ``Differential privacy: A survey of results,'' in \emph{International
  conference on theory and applications of models of computation}.\hskip 1em
  plus 0.5em minus 0.4em\relax Springer, 2008, pp. 1--19.

\bibitem{wei2020federated}
K.~Wei, J.~Li, M.~Ding, C.~Ma, H.~H. Yang, F.~Farokhi, S.~Jin, T.~Q. Quek, and
  H.~V. Poor, ``Federated learning with differential privacy: Algorithms and
  performance analysis,'' \emph{IEEE Transactions on Information Forensics and
  Security}, vol.~15, pp. 3454--3469, 2020.

\bibitem{stdp}
\BIBentryALTinterwordspacing
N.~Caporale and Y.~Dan, ``Spike timing–dependent plasticity: A hebbian
  learning rule,'' \emph{Annual Review of Neuroscience}, vol.~31, no.~1, pp.
  25--46, 2008, pMID: 18275283. [Online]. Available:
  \url{https://doi.org/10.1146/annurev.neuro.31.060407.125639}
\BIBentrySTDinterwordspacing

\bibitem{diehl2015fast}
P.~U. Diehl, D.~Neil, J.~Binas, M.~Cook, S.-C. Liu, and M.~Pfeiffer,
  ``Fast-classifying, high-accuracy spiking deep networks through weight and
  threshold balancing,'' in \emph{2015 International Joint Conference on Neural
  Networks (IJCNN)}.\hskip 1em plus 0.5em minus 0.4em\relax ieee, 2015, pp.
  1--8.

\bibitem{diehl2016conversion}
P.~U. Diehl, G.~Zarrella, A.~Cassidy, B.~U. Pedroni, and E.~Neftci,
  ``Conversion of artificial recurrent neural networks to spiking neural
  networks for low-power neuromorphic hardware,'' in \emph{2016 IEEE
  International Conference on Rebooting Computing (ICRC)}.\hskip 1em plus 0.5em
  minus 0.4em\relax IEEE, 2016, pp. 1--8.

\bibitem{neftci2019surrogate}
E.~O. Neftci, H.~Mostafa, and F.~Zenke, ``Surrogate gradient learning in
  spiking neural networks,'' 2019.

\bibitem{bntt}
Y.~Kim and P.~Panda, ``Revisiting batch normalization for training low-latency
  deep spiking neural networks from scratch,'' 2020.

\bibitem{adversarial_snn}
S.~Sharmin, N.~Rathi, P.~Panda, and K.~Roy, ``Inherent adversarial robustness
  of deep spiking neural networks: Effects of discrete input encoding and
  non-linear activations,'' 2020.

\bibitem{kim2021visual}
Y.~Kim and P.~Panda, ``Visual explanations from spiking neural networks using
  interspike intervals,'' \emph{arXiv preprint arXiv:2103.14441}, 2021.

\bibitem{mothukuri2021survey}
V.~Mothukuri, R.~M. Parizi, S.~Pouriyeh, Y.~Huang, A.~Dehghantanha, and
  G.~Srivastava, ``A survey on security and privacy of federated learning,''
  \emph{Future Generation Computer Systems}, vol. 115, pp. 619--640, 2021.

\bibitem{skatchkovsky2020federated}
N.~Skatchkovsky, H.~Jang, and O.~Simeone, ``Federated neuromorphic learning of
  spiking neural networks for low-power edge intelligence,'' in \emph{ICASSP
  2020-2020 IEEE International Conference on Acoustics, Speech and Signal
  Processing (ICASSP)}.\hskip 1em plus 0.5em minus 0.4em\relax IEEE, 2020, pp.
  8524--8528.

\bibitem{FedProx}
\BIBentryALTinterwordspacing
A.~K. Sahu, T.~Li, M.~Sanjabi, M.~Zaheer, A.~Talwalkar, and V.~Smith, ``On the
  convergence of federated optimization in heterogeneous networks,''
  \emph{CoRR}, vol. abs/1812.06127, 2018. [Online]. Available:
  \url{http://arxiv.org/abs/1812.06127}
\BIBentrySTDinterwordspacing

\bibitem{fedma}
H.~Wang, M.~Yurochkin, Y.~Sun, D.~Papailiopoulos, and Y.~Khazaeni, ``Federated
  learning with matched averaging,'' \emph{arXiv preprint arXiv:2002.06440},
  2020.

\bibitem{discrete_lif}
\BIBentryALTinterwordspacing
Y.~Wu, L.~Deng, G.~Li, J.~Zhu, and L.~Shi, ``Spatio-temporal backpropagation
  for training high-performance spiking neural networks,'' \emph{Frontiers in
  Neuroscience}, vol.~12, p. 331, 2018. [Online]. Available:
  \url{https://www.frontiersin.org/article/10.3389/fnins.2018.00331}
\BIBentrySTDinterwordspacing

\bibitem{rathi2020enabling}
N.~Rathi, G.~Srinivasan, P.~Panda, and K.~Roy, ``Enabling deep spiking neural
  networks with hybrid conversion and spike timing dependent backpropagation,''
  2020.

\bibitem{snn_bp}
S.~M. Bohte, ``Error-backpropagation in networks of fractionally predictive
  spiking neurons,'' in \emph{Artificial Neural Networks and Machine Learning
  -- ICANN 2011}, T.~Honkela, W.~Duch, M.~Girolami, and S.~Kaski, Eds.\hskip
  1em plus 0.5em minus 0.4em\relax Berlin, Heidelberg: Springer Berlin
  Heidelberg, 2011, pp. 60--68.

\bibitem{optimal_grad_compression}
A.~Albasyoni, M.~Safaryan, L.~Condat, and P.~Richtárik, ``Optimal gradient
  compression for distributed and federated learning,'' 2020.

\bibitem{vgg}
K.~Simonyan and A.~Zisserman, ``Very deep convolutional networks for
  large-scale image recognition,'' \emph{arXiv preprint arXiv:1409.1556}, 2014.

\bibitem{CIFAR}
A.~Krizhevsky \emph{et~al.}, ``Learning multiple layers of features from tiny
  images,'' 2009.

\bibitem{bayesianfl}
M.~Yurochkin, M.~Agarwal, S.~Ghosh, K.~Greenewald, T.~N. Hoang, and
  Y.~Khazaeni, ``Bayesian nonparametric federated learning of neural
  networks,'' \emph{arXiv preprint arXiv:1905.12022}, 2019.

\bibitem{horowitz20141}
M.~Horowitz, ``1.1 computing's energy problem (and what we can do about it),''
  in \emph{2014 IEEE International Solid-State Circuits Conference Digest of
  Technical Papers (ISSCC)}.\hskip 1em plus 0.5em minus 0.4em\relax IEEE, 2014,
  pp. 10--14.

\end{thebibliography}

\begin{IEEEbiography}[{\includegraphics[width=1in,height=1.25in,clip,keepaspectratio]{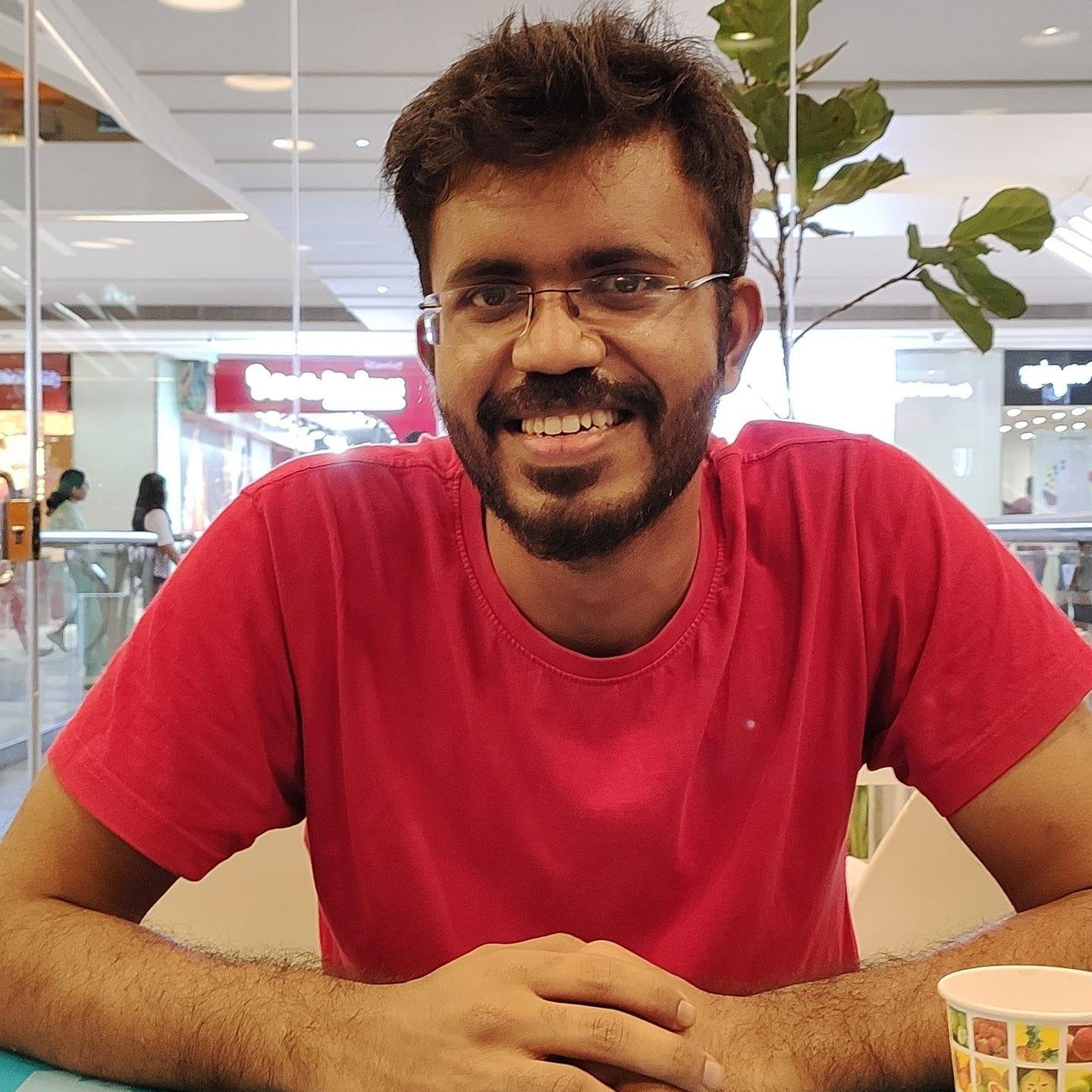}}]{Yeshwanth Venkatesha} received B.Tech.\ in Computer Science and Engineering from Indian Institute of Technology and Science Kharagpur, India, in 2017. He joined Yale University, USA, in 2020 as a Ph.D. student in the Electrical Engineering department. Prior to joining Yale University, he worked as a Data Scientist and Software Engineer at Walmart Labs India and Samsung Advanced Institute of Technology India respectively. His research interests lie in the areas of efficient processing of neural networks and distributed machine learning.
\end{IEEEbiography}

\begin{IEEEbiography}[{\includegraphics[width=1in,height=1.23in,clip,keepaspectratio]{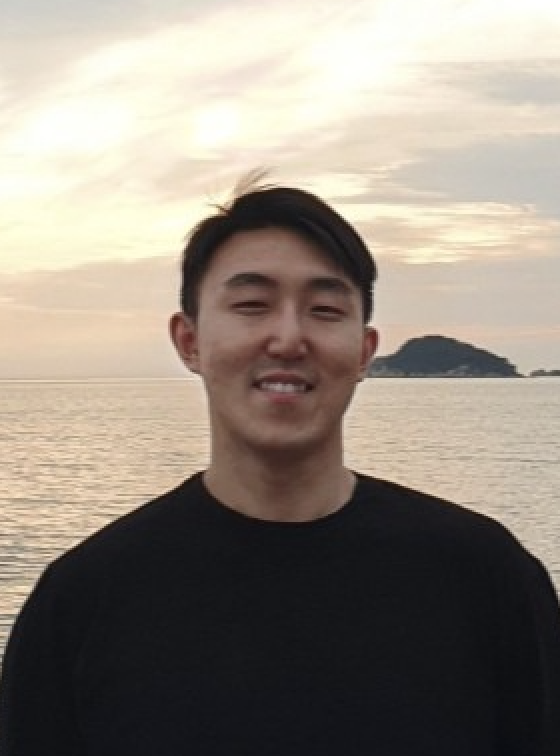}}]{Youngeun Kim}
is currently working toward the Ph.D. degree in Electrical Engineering at Yale University, New Haven, CT, USA.
Prior to joining Yale,  he worked as a full-time student internship at T-Brain, AI Center, SK telecom, South Korea. 
He received his B.S. degree in Electronic Engineering from Sogang University, South Korea, in 2018 and M.S. degree in Electrical Engineering from Korea Advanced Institute of Science and Technology (KAIST), in 2020. 
His research interests include neuromorphic  computing, computer vision, and deep learning.
\end{IEEEbiography}

\begin{IEEEbiography}[{\includegraphics[width=1in,height=1.25in,clip,keepaspectratio]{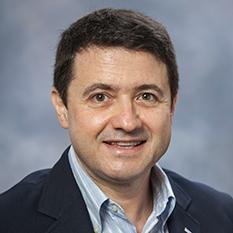}}]{Leandros Tassiulas} received
the Ph.D. degree in electrical engineering from the
University of Maryland, College Park, MD, USA,
in 1991.
He held faculty positions at Polytechnic University, New York, NY, USA, the University of
Maryland, and the University of Thessaly, Greece.
He is currently the John C. Malone Professor of
electrical engineering with Yale University. His
research interests include computer and communication networks, with an emphasis on fundamental mathematical models
and algorithms of complex networks, architectures and protocols of wireless
systems, sensor networks, novel internet architectures, and experimental
platforms for network research. His most notable contributions include the
max-weight scheduling algorithm and the back-pressure network control
policy, opportunistic scheduling in wireless, the maximum lifetime approach
for wireless network energy management, and the consideration of joint
access control and antenna transmission management in multiple antenna
wireless systems.
Dr. Tassiulas’s research has been recognized by several awards, including
the IEEE Koji Kobayashi Computer and Communications Award, the Inaugural INFOCOM 2007 Achievement Award for fundamental contributions
to resource allocation in communication networks, the INFOCOM 1994 and
2017 Best Paper Awards, the National Science Foundation (NSF) Research
Initiation Award in 1992, the NSF CAREER Award in 1995, the Office
of Naval Research Young Investigator Award in 1997, and the Bodossaki
Foundation Award in 1999.

\end{IEEEbiography}

\begin{IEEEbiography}[{\includegraphics[width=1in,height=1.25in,clip,keepaspectratio]{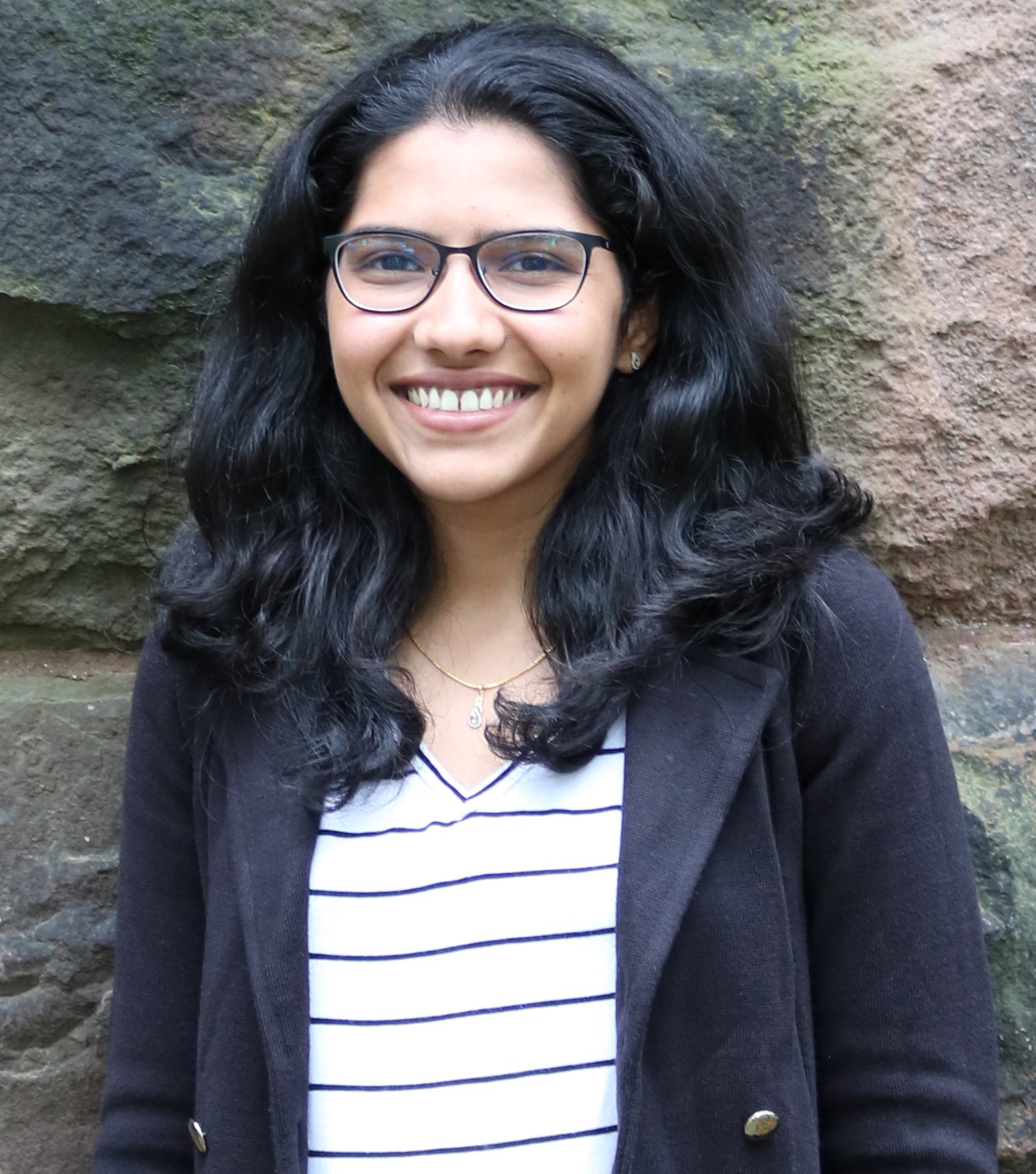}}]{Priyadarshini Panda} obtained her Ph.D. from Purdue University, USA, in 2019. She joined Yale University, USA, as an assistant professor in the Electrical Engineering department in August, 2019. She received the B.E. degree in Electrical \& Electronics Engineering and the M.Sc. degree in Physics from B.I.T.S. Pilani, India, in 2013. She was the recipient of outstanding student award in physics for academic excellence. From 2013-14, she worked in Intel, India on RTL design for graphics power management. She has also worked with Intel Labs, USA, in 2017 and Nvidia, India in 2013 as research intern. She has been awarded the Amazon Research Award for 2019-20. Her research interests lie in robust and energy efficient neuromorphic computing, deep learning and related applications through algorithm-hardware co-design.

\end{IEEEbiography}



%

\end{document}